%% file: main.tex
\documentclass{article}

\usepackage{arxiv}

\usepackage[utf8]{inputenc} 

\usepackage{amsfonts}       
\usepackage{nicefrac}       
\usepackage{microtype}      
\usepackage{lipsum}

\usepackage{color}
\usepackage{listings}
\usepackage{booktabs}
\usepackage{array}
\usepackage{graphicx}
\usepackage{longtable}

\usepackage{cite}
\usepackage{url}
\usepackage{fancyhdr}

\usepackage{soul}

\usepackage{float}
\usepackage{listings}
\usepackage{mdwmath}
\usepackage{mdwtab}
\usepackage{multirow}
\usepackage{multicol}
\usepackage{rotating}
\usepackage{setspace}
\usepackage[utf8]{inputenc}
\usepackage{lineno}
\usepackage{listings}

\usepackage{mdwmath}
\usepackage{mdwtab}
\usepackage{multirow}
\usepackage{multicol}
\usepackage{array}
\usepackage{booktabs}

\usepackage{enumitem}
\usepackage{xspace}
\usepackage[export]{adjustbox}
\usepackage{graphicx}
\usepackage{color,soul}
\usepackage{rotating}
\usepackage{setspace}
\usepackage{amsmath} 
\usepackage{amssymb}
\usepackage{float}
\usepackage{xcolor}

\usepackage{hyperref}
\usepackage[numbers]{natbib}
\usepackage{url}

\title{Software Engineering for Responsible AI:\\An Empirical Study and Operationalised Patterns}

\author{Qinghua Lu, Liming Zhu, Xiwei Xu, Jon Whittle, David Douglas, Conrad Sanderson\\
Data61, CSIRO, Australia\\
{firstname.lastname}@csiro.au}

\begin{document}

\maketitle

\begin{abstract}
Although artificial intelligence (AI) is solving real-world challenges and transforming industries, there are serious concerns about its ability to behave and make decisions in a responsible way. Many AI ethics principles and guidelines for responsible AI have been recently issued by governments, organisations, and enterprises. However, these AI ethics principles and guidelines are typically high-level and do not provide concrete guidance on how to design and develop responsible AI systems. To address this shortcoming, we first present an empirical study where we interviewed 21 scientists and engineers to understand the practitioners' perceptions on AI ethics principles and their implementation. We then propose a template that enables AI ethics principles to be operationalised in the form of concrete patterns and suggest a list of patterns using the newly created template. These patterns provide concrete, operationalised guidance that facilitate the development of responsible AI systems.

\end{abstract}

artificial intelligence, AI, machine learning, responsible AI, ethics, software engineering, software architecture, DevOps

\section{Introduction}
Artificial intelligence (AI) continues demonstrating its positive impact on society and successful adoptions in data rich domains. The global AI market was valued at approx.~USD 62 billion in 2020 and is expected to grow with an annual growth rate of 40\% from 2021 to 2028~\cite{Artificial_Intelligence_Market_Size}. Although AI is solving real-world challenges and transforming industries, there are serious concerns about its ability to behave and make decisions in a responsible way. 

\begin{figure*}
\centering
\includegraphics[width=0.8\textwidth,height=0.40\textwidth]{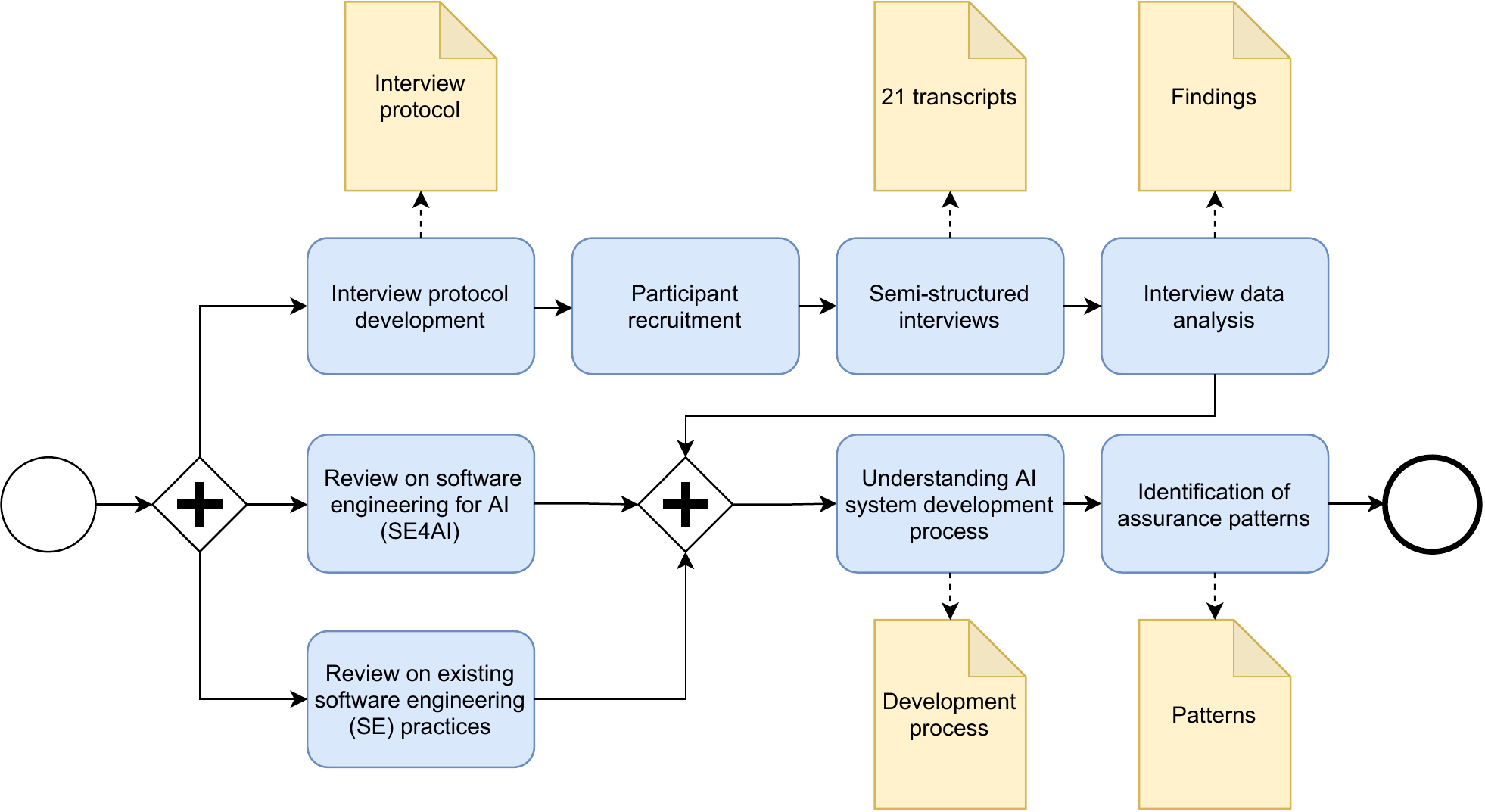}
\vspace{-1ex}
\caption{Overview of the employed methodology.} \label{fig:methodology}
\end{figure*}

To achieve responsible AI, both ethical and legal aspects may need to be considered. As law is usually considered to set the minimum standards of behaviour while ethics establishes the maximum standards, 
throughout this paper we use the terms \textit{responsible AI}, \textit{ethical AI} and \textit{ethics} to cover the broader set of requirements. Trustworthy AI refers to AI systems that embody the responsible AI principles and requirements~\cite{oecd}. Many AI ethics principles and guidelines for responsible AI have been recently
issued by governments, organisations, and enterprises ~\cite{jobin2019global,fjeld2020principled}. However, these principles are typically high-level and do not provide concrete guidance on how to develop responsible AI systems. 
To address this, we pose the research questions listed below.

\textbf{RQ1: What are the current states and potential challenges developers are facing in dealing with responsible AI issues during the development of AI systems?}
We perform an empirical study where we interviewed 21 AI scientists and engineers with various backgrounds and expertise. We asked participants what ethical issues they have considered in their AI projects and how the ethical issue were addressed or they envisioned can be addressed. Based on the interview results, we reveal several major findings:
\textbf{(1)}~The current approach is often a done-once-and-forget type of risk assessment at a particular development step, which is not sufficient for the highly uncertain and continual learning AI systems;
\textbf{(2)}~Responsible AI requirements are either omitted or mostly stated as high-level objectives, and not specified explicitly in verifiable way as system outputs or outcomes;
\textbf{(3)}~Although responsible AI requirements have the characteristics of cross-cutting quality and non-functional requirements amenable to architecture and design analysis, system-level architecture and design are under-explored;
\textbf{(4)}~There is a strong desire for continuous monitoring and validation of AI systems post deployment for responsible AI requirements, where current MLOps/AIOps practices provide limited guidance.

\textbf{RQ2: How can AI ethics principles be operationalised into concrete practice that AI developers can use throughout the lifecycle of AI systems?}
We design a pattern template that enables AI ethics principles to be operationalised in the form of concrete patterns. We then suggest a list of process and design patterns using the newly created template throughout the lifecycle of an AI system based on the results of the interviews, literature review, as well as existing software development and design practices.



The major contributions of our study are as follows:
\begin{itemize}[leftmargin=*]
\setlength{\itemsep}{0pt}
\setlength{\parsep}{0pt}
\setlength{\parskip}{0pt}
\item To the best of our knowledge, this is the first in-depth study that explores practitioners' perceptions on AI ethics principles and their implementation. 

\item We identify the AI \textit{system} development and operation (AIOps/MLOps) process that integrates with the AI \textit{model} development process that includes data collection, feature engineering, model training, evaluation and updates.

\item We propose a template to define patterns for operationalising responsible AI and summarise a list of ethical assurance patterns using the newly designed template throughout the lifecycle of an AI system. The patterns provide a concrete, operationalised guidance that can be easily applied and extended by AI developers to develop responsible AI systems.
\end{itemize}

\noindent
We continue the paper as follows.
Section~\ref{methodology} overviews the methodology.
Section~\ref{process} identifies the development process.
Section~\ref{findings} discusses the findings.
Section~\ref{mechanisms} suggests a list of patterns.
Section~\ref{validity} discusses the threats to validity.
Section~\ref{related_work} covers related work.
Concluding remarks are given in Section \ref{conclusion}.

\section{Methodology}
\label{methodology}

An overview of the methodology is given in Fig.~\ref{fig:methodology}.
The major findings were extracted through interviews,
while the AI system development process and ethical assurance patterns were identified based on the interview results,
literature review on software engineering for AI and machine learning (SE4AI/SE4ML),
and existing software engineering (SE) practices. 

The interviewees were from a research institute 
and sought via ``call for participation'' emails
as well as via follow-up recommendations given by the interviewees,
until a saturation of perspectives were reached.
21 interviews were conducted from February to April 2021.
The interviewees are from various backgrounds,
with a large variation in the interviewees' degree of experience and responsibility. 
10~interviewees worked primarily in computer science,
6~interviewees worked in the health \& biosecurity area,
and 5~interviewees worked in the land \& water area.
The job positions of the interviewees included:
postgraduate student (1),
research scientist (1),
senior research scientist (4),
principal research scientist (2),
principal research engineer (1),
team leader (8),
group leader (4).
The gender split was approximately 24\% females and 76\% males.

\begin{figure}[b!]
\vspace{-1ex}
\centering
\fbox{%
\centering
\begin{minipage}{0.96\columnwidth}
\small
\begin{enumerate}[leftmargin=*]
\item
{\bf Privacy Protection \& Security.}
AI systems should respect and uphold privacy rights and data protection, and ensure the security of data.

\item
{\bf Reliability \& Safety.}
AI systems should reliably operate in accordance with their intended purpose throughout their lifecycle.

\item
{\bf Transparency \& Explainability.}
\textit{Transparency:} there should be transparency and responsible disclosure to ensure people know when they are being
significantly impacted by an AI system, and can find out when an AI system is engaging with them.
\textit{Explainability:}
what the AI system is doing and why, such as the system's processes and input data.

\item
{\bf Fairness.}
AI systems should be inclusive and accessible, and should not involve or result in unfair discrimination
against individuals, communities or groups.

\item
{\bf Contestability.}
When an AI system significantly impacts a person, community, group or environment, there should be a
timely process to allow people to challenge the use or output of the system.

\item
{\bf Accountability.}
Those responsible for the various phases of the AI system lifecycle should be identifiable and accountable
for the outcomes of the system, and human oversight of AI systems should be enabled.

\item
{\bf Human-centred Values.}
AI systems should respect human rights, diversity, and the autonomy of individuals.

\item
{\bf Human, Social \& Environmental Wellbeing.}
AI systems should benefit individuals, society and the environment.
\end{enumerate}
\end{minipage}
} 
\caption
  {
  An adapted summary of 8 voluntary high-level ethics principles for AI,
  as promulgated by the Australian Government~\cite{DISER_2020}.
  }
\label{fig:principles_summary}
\end{figure}


\begin{table}[t!]
\normalsize
\centering
\caption{Incidence of themes related to AI ethics principles.}
\label{tab:incidence}
\begin{tabular}{p{0.65\columnwidth}p{0.25\columnwidth}}
\toprule
\bf{Principle} &
\bf{Incidence}  \\
\midrule

Privacy Protection \& Security & 17 / 21  (81\%) \\
\cmidrule{1-2}

Reliability \& Safety & 19 / 21  (90\%) \\
\cmidrule{1-2}

Transparency \& Explainability & 18 / 21  (86\%) \\
\cmidrule{1-2}

Accountability & 13 / 21  (62\%) \\
\cmidrule{1-2}

Contestability & 8\; / 21  (38\%) \\
\cmidrule{1-2}

Fairness & 10 / 21  (47\%) \\
\cmidrule{1-2}

Human-Centred Values & 3\; / 21  (14\%) \\
\cmidrule{1-2}

Human, Societal and Environmental (HSE) Wellbeing & 11 / 21  (52\%) \\

\bottomrule
\end{tabular}
\vspace{-2ex}
\end{table}

\begin{figure*}
\centering
\includegraphics[width=0.99\textwidth]{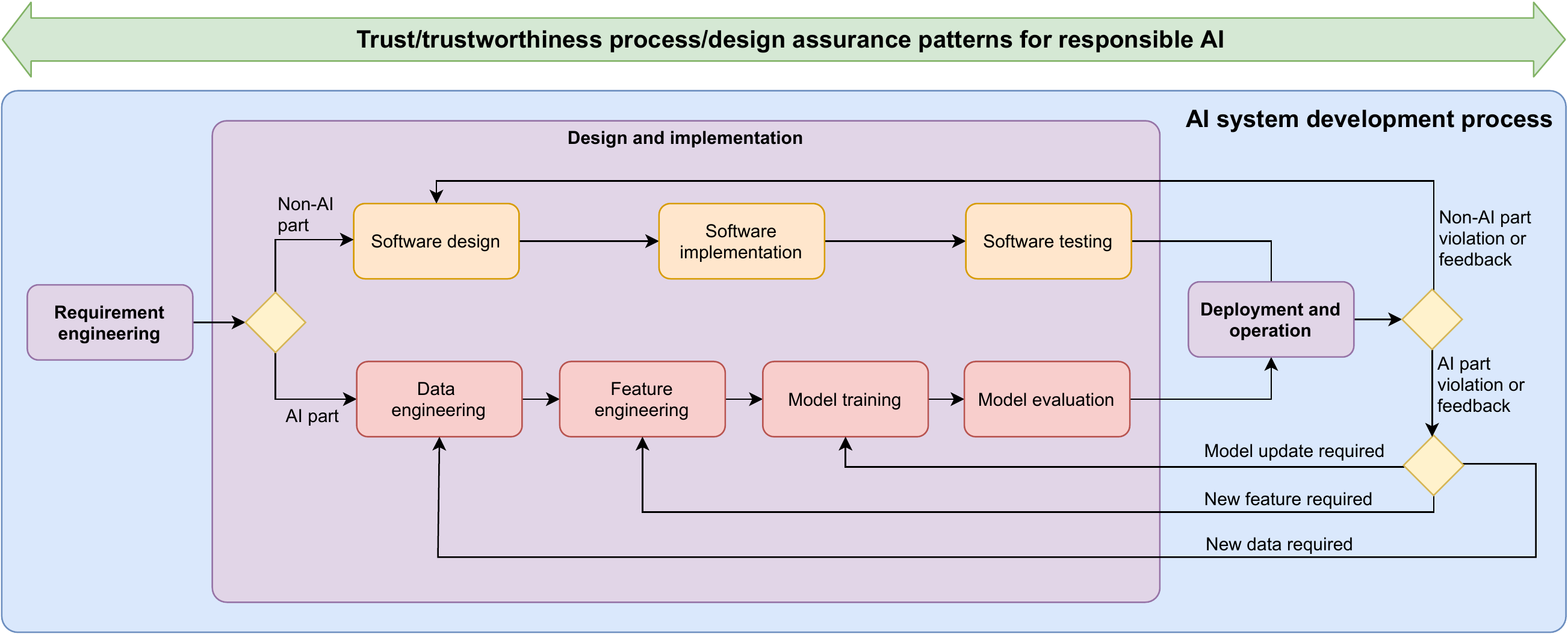}
\caption{AI system development process.} \label{process_overview}
\end{figure*}

The interviews were conducted by three project team members with various research backgrounds (machine learning, software engineering, ethics in AI, respectively), in a face-to-face setting and/or via video teleconferencing.
Prior to each interview, each interviewee was given a summary of Australia's AI ethics principles~\cite{DISER_2020}
(as shown in Fig.~\ref{fig:principles_summary}),
to ensure all interviewees are aware of the principles.
The interviews ranged from approximately 22 to 59 minutes in length, with a median length of approximately 37 minutes.
We followed the methodology employed in \cite{interview_methodology_2018} to stop interviews when saturation of findings was reached.

The transcripts were analysed using theoretical thematic analysis~\cite{thematic_analysis}.
This analysis used a theoretical approach to coding the interview data by using the eight AI ethics principles as themes.
Concepts identified in discussions of specific principles were recorded as sub-themes related to that principle.
We summarised the findings based on the interview analysis data. Table \ref{tab:incidence} shows the incidence of themes related to AI ethics principles across the interviews.
The top three principles covered in the interviews are \textit{Reliability \& Safety}, \textit{Transparency \& Explainability}, and \textit{Privacy Protection \& Security}.
Principles which were covered in roughly half the interviews are \textit{Accountability}, \textit{HSE Wellbeing}.
The \textit{Human-Centred Values} principle was covered the least in the interviews.

\section{AI System Development Process}
\label{process}
Fig.~\ref{process_overview} illustrates an overview of AI development process.
The process starts with requirement analysis.
In this phase, we need to identify the requirements and constraints placed by stakeholders.
In recent years, responsible software, responsible technology and human values in software
has become an important field of study~\cite{Whittle19}.
Responsible/ethical AI (system) is a sub-field within the responsible technology (software) field.
However, compared with traditional software, AI systems also need to consider
requirements about models, training data, system autonomy oversight
and may emphasise certain ethical requirements more
due to AI-based autonomous and potentially opaque behaviour and decision making.

Once the requirements are identified, the process is divided
into two sub-process for non-AI part and AI part, respectively.
The non-AI part sub-process includes design, implementation, and testing of non-AI components.
The AI part sub-process is the AI development process for model production,
which covers data engineering, feature engineering, model training, model evaluation and updates.
The converged phase for non-AI part and AI part is the deployment and operation of the AI system.
Some key differences in the deployment and operation of AI systems
are often the continual learning of AI components based on new data,
the higher degree of uncertainty and risks associated with the autonomy of the AI component,
and validation of outcomes
(i.e. did the system provide the intended benefits and behave appropriately given the situation?)
rather than just outputs (e.g. precision, accuracy and recall)~\cite{cmu_human_centered_ai}.

\section{Findings}
\label{findings}
In this section, we report our findings for each of the categories that were identified using open card sorting on interview contents. For each category, we select the most meaningful comments and highlight our observations. 

\subsection{Overall development process}

\noindent \textbf{Ethical risk assessment.} 
Understanding and managing risk is particularly important for AI systems as they may be highly uncertain
and may involve continual learning.
We found some ethical risk assessment frameworks were used in practice.
One interviewee stated \textit{``There was a privacy impact assessment. We went through a lengthy process to understand the privacy concerns and build in provisions to enable privacy controls and people to highlight things that they didn’t want to be visible'' (P10).}
However, such kind of approach is a done-once-and-forget type of risk assessment
and not sufficient for AI systems that are highly uncertain and continually learn.
Furthermore, various practitioners approach risk differently.
One interviewee suggested fail-safe by design should be considered and noted that
\textit{``there’s only so much you can think ahead about what those failure modes might be'' (P16).}
One interviewee argued \textit{``Once I know that it works most of the time I don’t need explainability,
I don’t need transparency. It’s just temporary to establish the risk profile''~(P11).}

\begin{figure}[h!]
\centering
\noindent
\fbox{%
\begin{minipage}{0.99\columnwidth}
\normalsize
    \textbf{Finding 1:} The current practice is a done-once-and-forget type of risk assessment at a particular development step, which is not sufficient for the highly uncertain and continual learning AI systems.
\end{minipage}}
\end{figure}

\noindent \textbf{Trust vs. trustworthiness.}
Trustworthiness is the ability of an AI system to meet AI ethics principles, while trust is users' subjective estimates of the trustworthiness of the AI system \cite{zhu2021ai}. Even for a highly trustworthy AI system, gaining the trust from humans is another challenge that must be addressed carefully for the AI system to be widely accepted. This is because a user's subjective estimates of the AI system's trustworthiness may have a significant gap compared to the AI system's inherent trustworthiness. It can also be the other way around when a user overestimates a system's trustworthiness and put excessive trust into it.
We found many interviewees have recognised the importance of human trust in AI. One interviewee stated: \textit{``A lot of the work that we do trust comes as an important factor here, that a user or somebody who takes that information, wants to be able to trust it'' (P9).} One of the obstacles for the development of AI systems is gaining and maintaining the trust from the data providers. One interviewee noted \textit{``you build the trust with the data providers, so more people can give you data and increase your data representability'' (P2).} 
One interviewee pointed out evidence need to be offered to drive trust: \textit{``Because you justifiably want to trust that system and not only ask people do you trust it? I mean they need some evidence. You can build this into your system to some degree. So that's very important'' (P12).}



\begin{figure}[h!]
\begin{center}
\noindent
\fbox{%
\begin{minipage}{0.99\columnwidth}
\normalsize
    \textbf{Finding 2:} The inherent trustworthiness of an AI system for various ethical principles and the perceived trust of the system are often mixed in practice. Even for a highly trustworthy AI system, gaining the trust from humans is a challenge that must be addressed carefully for the AI system to be widely accepted. Process and product mechanisms can be leveraged to achieve trustworthiness for various ethical principles, whereas process and product evidence need to be offered to drive trust. 
\end{minipage}}
\end{center} 
\end{figure}

\noindent \textbf{Ethical credentials.} 
AI industry requires responsible AI components and products at each step of the value chain. AI system vendors often supply products by assembling commercial or open-source AI and/or non-AI components. Some interviewees agreed credential schemes can enable responsible AI by attaching ethical credentials to AI components and products. One interviewee commented \textit{``Getting those certificates, it always helps. As long as there is standardisation around it.'' (P13).} There have been certificates for the underlying hardware of AI systems. One interviewee pointed out \textit{``A lot of hardware is actually certified. I mean in (...) full size aviation. you have at least a certification. So when you buy something you get some sort of guarantees'' (P12).}

\begin{figure}[h!]
\begin{center}
\noindent
\fbox{%
\begin{minipage}{0.99\columnwidth}
\normalsize
    \textbf{Finding 3:} Human trust in AI can be improved by attaching ethical credentials to AI components/products since the vendors often supply products by assembling commercial or open-source AI or non-AI components. 
\end{minipage}}
\end{center} 
\end{figure}

\noindent \textbf{Requirement-driven development vs. outcome-driven development.} 
We observed there are two forms of development mentioned by the interviewees: requirement-driven development and outcome-driven development~\cite{jan2019}. Among the ethical requirements/principles, privacy and security is one of the most discussed requirements. 
One interviewee noted privacy requirements: \textit{``To protect those privacy and de-identification requirements, you’ll be aggregating so that people can’t be uniquely identified'' (P1).} In relation to outcome-driven development, one interviewee emphasised the development is a continual process: \textit{``This is a continual and iteration process, human need to continually to evaluate the performance, identify the gap and provide insight into what’s missing. Then go back to connect data and refine the model'' (P2).} 

\begin{figure}[h!]
\begin{center}
\noindent
\fbox{%
\begin{minipage}{0.99\columnwidth}
\normalsize
    \textbf{Finding 4:} Developing AI systems requires seamless integration of requirement-driven development and outcome-driven development.
\end{minipage}}
\end{center} 
\end{figure}



\noindent \textbf{End-to-end system-level development tools.}
An AI system consists of AI components and non-AI components
that are interconnected and work together to achieve the system's objective.
An AI model needs to be integrated with the system to perform the required functions.
Combining AI and non-AI components may create new emergent behavior and dynamics.
Therefore, ethics need to be considered at system-level,
including AI components, non-AI components and their connections.
For example, the effect of actions decided by the AI model could be collected
through the feedback component built into the AI system.
Although most of the interviewees are research scientists/engineers
who mainly worked on research projects and focused on model development,
some of them did recognise the significance of system-level thinking in AI projects.
One interviewee commented
\textit{``Well, it's just that the design ways in which that AI was designed and deployed as an end-to-end solution, it wasn't that AI sat in the middle, right? It actually had to sit within the system'' (P14).}
We also found that the management of AI ethics principles was heavily relied on manual practice.
One interviewee pointed out \textit{``We had to go through a lot of data and make sure that there was not a single frame with a person in it'' (P13).}
This accidental collection of sensitive data issue could be addressed automatically
using AI enabled human detection tools. 

\begin{figure}[h!]
\begin{center}
\noindent
\fbox{%
\begin{minipage}{0.99\columnwidth}
\normalsize
    \textbf{Finding 5:} An AI model needs to be integrated with the system to perform the required functions. Combining AI and non-AI components create new emergent behaviour and dynamics, which require system-level ethical consideration. Implementation of AI ethics principles still heavily relied on manual operations. There is lack of end-to-end development tools to support continuous assurance of AI ethics.
\end{minipage}}
\end{center} 
\end{figure}


\subsection{Requirement engineering}

\noindent \textbf{Ethical requirements.} We found some ethics principles, such as human, societal, environmental well-being, were sometimes omitted and stated only as a project objective rather than verifiable requirements or outcomes. One interviewee stated \textit{``People are presented with a clear project objective upfront, and the project leader might frame the project with we're working on improving [a grass specie] yield forecasting using machine learning. You do feel good about working on projects that provide environmental benefit'' (P9).} Ethical AI requirements need to be analysed, verified and validated by a wide range of experts beyond the software developers (such as hardware engineers, culture expert, end users). 



\begin{figure}[h!]
\begin{center}
\noindent
\fbox{%
\begin{minipage}{0.99\columnwidth}
\normalsize
    \textbf{Finding 6:} Responsible AI requirements are either omitted or mostly stated as high-level objectives, and not specified explicitly in a verifiable way as expected system outputs (to be verified/validated) and outcomes (e.g. benefits). Requirements engineering methods need to be extended with ethical aspects for AI systems.
\end{minipage}}
\end{center} 
\end{figure}

\noindent \textbf{Scope of responsibility.}
Based on our interview results, we found that there were various perceptions on the meaning of responsible AI. One interviewee raised a question of meaning of responsibility in the context of autonomous drone systems \textit{``The question is what happens if [the] remote pilot is really there, flicks the switch [to disable the system] and the system doesn’t react? The remote pilot is not always in full control of [the drone] because of technical reasons [such as a failed radio link]'' (P12).}
The various meanings and interpretations of the word ``responsible'' have already received considerable attention. Tigard et al.~\cite{Tigard2021} introduce three varieties of responsibility,
including the normative interpretation (i.e. behaving in positive, desirable and socially acceptable ways),
the possessive interpretation (i.e. having a duty and obligation) and descriptive interpretation
(i.e. worthy of a response - answerable).
Lima et al.~\cite{responsibility_meaning_chi21} summarise eight meanings of responsibility.
We observe interviewees touched on all the three varieties of Tigard's meanings~\cite{Tigard2021}
and considered all of them as important.
Furthermore, timeliness needs to be considered for responsibility.
One interviewee commented \textit{``whether the stuff works in 10 years, it’s not under our control (...) and we shouldn’t really care about it'' (P11).}

\begin{figure}[h!]
\begin{center}
\noindent
\fbox{%
\begin{minipage}{0.99\columnwidth}
\normalsize
    \textbf{Finding 7:} The various meanings and interpretations of the word ``responsible'' have already received considerable attention. There are three varieties of responsibility including the normal interpretation, the possessive interpretation, and descriptive interpretation. 
\end{minipage}}
\end{center} 
\end{figure}

\subsection{Design and implementation}

\noindent \textbf{AI in design.}
AI is an uncertain and complex technology which is often hard to explain thus making detailed risk assessment difficult. 
 One interviewee commented \textit{``When do you have a complete assessment really? Especially with systems that change over time and based on sensory input. [...] It’s very difficult'' (P12).} 
Adopting AI or not can be considered as a major architectural design decision when designing a software system. An architect can also design an AI component that can be switched off during run-time or changed from decision mode to suggestion mode. It is necessary to let humans make judgement throughout the lifecycle of an AI system, e.g. whether to adopt AI in design or whether to accept the recommendations made by the AI systems. One interviewee explained overriding the recommended decisions with an clinical system example: \textit{``there was actually a defined process where if a patient was not flagged as being high risk, [...] clinicians were still allowed to include the patient into the next step clinical review'' (P18).}

\begin{figure}[h!]
\centering
\noindent
\fbox{%
\begin{minipage}{0.99\columnwidth}
\normalsize
    \textbf{Finding 8:}
    AI is an uncertain and complex technology which is often hard to explain,
    thus making detailed risk assessment difficult.
    Adopting AI can be considered as a major architectural design decision when designing a software system.
    Furthermore, an AI component can be designed to be flexibly switched off at run-time
    or changed from decision mode to suggestion mode.
\end{minipage}}
\end{figure}

\noindent \textbf{Trade-offs between ethical principles in design.}
Several interviewees pointed out there are trade-offs between ethics principles (e.g. privacy vs. reliability/accountability, fairness vs. reliability). 
One interviewee commented \textit{``
If you’ve got other ways of protecting privacy that don’t involve aggregating, then you can be actually getting better distributional properties''(P01).} 
However, there was not much discussion about the methods on how to deal with the trade-offs. 


The reliability of AI depends on the quantity and quality of the training data. One interview noted that \textit{``if you’re training a model without a lot of data, you can actually get some really weird results'' (P9).} Obtaining a sufficient number of samples can be challenging, as obtaining one sample can be high in terms of both financial/time cost and privacy issues in domains such as genomics (P3). Federated learning was mentioned to deal with privacy and security ethical concerns in addition to the data hungriness issues \textit{``different research institutions from around the world can collaborate, because they don’t have to give up their data. They don’t have to share their data'' (P3).} There was a desire to use such architecture styles and specific patterns to handle some ethical AI requirements.

\begin{figure}[h!]
\begin{center}
\noindent
\fbox{%
\begin{minipage}{0.99\columnwidth}
\normalsize
    \textbf{Finding 9:} There are trade-offs between some AI ethics principles. The current practice to deal with the trade-offs is usually the developers following one principle while overwriting the other rather than building balanced trade-offs with stakeholders making the ultimate value and risk call.
\end{minipage}}
\end{center} 
\end{figure}

\noindent \textbf{Design process for ethics.}
We found the reuse of models and other AI pipeline components is desired since training models and building various components in the model development pipeline is time-consuming and costly. There was extension to the reuse of the overall architecture and design of the AI system due to its dependency on the costly and complex pipeline. Similar issues were reported in literature regarding architecture degradation and accumulating of high technical debt over time \cite{google_technical_debt}.  However, the ethical AI consequence of the reuse was not well understood. One interviewee highlighted \textit{``What we have gone beyond the project we hope to achieve is we’ll have the whole pipeline in place. Once we have different data from a different environment that’s not associated to that particular company that they labelled and they recorded. We already have something in place that we can train with different data. As long as it’s not the same data - it’s a different type of environment - that’s fine'' (P13).} It would be helpful to develop modelling languages and representation tools to capture various ethical concerns and represent the AI components to stakeholders to improve explainability. The representation is not only about model structure, maybe depending on who developers work with, show various views and ethical concerns of AI components.

\begin{figure}[h!]
\begin{center}
\noindent
\fbox{%
\begin{minipage}{0.99\columnwidth}
\normalsize
    \textbf{Finding 10:} Although responsible AI requirements have the characteristics of cross-cutting quality and non-functional requirements amenable to architecture/design analysis and reusable patterns, they were under explored in the projects.
\end{minipage}}
\end{center} 
\end{figure}

\noindent \textbf{Design for explainability and interpretability.}
Explainability and interpretability are two emerging quality attributes for AI systems.
We found some interviewees have considered explainability and interpretability in practice and adopted human-centred approaches taking into account users' background, culture, and preferences to improve human trust. Explainability defines being able to come up with features in an interpretable domain that contribute to some explanation about how an outcome is achieved. The recommendations made by the AI systems are often not that useful to assist users to make decisions, unless the system shows the indicators and factors for why that prediction was given. One interviewee noted that \textit{``there have been instances where we’ve chosen an explainable model which has slightly lowered performance to a non-explainable model which has higher performance but would be harder to convey the reasoning behind the prediction'' (P18).} Interpretability is the ability of an AI system to provide a understandable description of a stimulus (e.g., model output) in stakeholders' familiar terms. One interviewee stated \textit{``I'm really experimenting now with how we actually show the data so that it can be interpreted by people? So we're playing around with data visualisation tools now to say how do we bring that data to bear and going out there and saying does this make sense to you? We designed all these reports which just show the data in different ways and part of that was - do you like the way this is going or is there things you'd like to see?'' (P14).} 



Most of the actions for explainability that were discussed by the interviewees were around the interface design of AI systems. 
One interviewee commented \textit{``That interface was really responsible behind - nobody seems to ask about, what's the predictive performance of the algorithm [in the initial stakeholder meeting]? It's around, can I look at your interface and look at - see a couple of patient risk profiles and then understand that.'' (P18).}

It is necessary to calibrate trust over time to match AI systems' trustworthiness. One interviewee stated \textit{``There is no need to explain anything if you know the risk and and if you have a long enough time to look over it. 
So this explainability thing, it’s just a temporary requirement until the risk is known'' (P14).} 

\begin{figure}[h!]
\begin{center}
\noindent
\fbox{%
\begin{minipage}{0.99\columnwidth}
\normalsize
    \textbf{Finding 11:} Human-centred approaches have been adopted for explainability and interpretability taking into account users' background and preferences to improve human trust in AI. 
\end{minipage}}
\end{center} 
\end{figure}

\subsection{Deployment and operation}
\noindent \textbf{Continuous validation of AI ethics.}
There is a strong desire for continuously monitoring and validating AI systems post deployment for ethical requirements. One interviewee commented: \textit{``It's up to us to come with technology that makes it acceptable for them to implement measurements in that respect and being able to prove compliance or even signal a trend like you're compliant now, but because we can see that your [values] are slowly going up and that's your threshold, so you're approaching it'' (P7).} 
Awareness of potential mismatches between training data and real-world data is necessary to prevent the trained model from being unsuitable for its intended purpose (P4). Model update or recalibration on new data were seen as important for the reliability of AI systems. The models may need to be retrained or recalibrated to properly take advantage of user feedback, newer and/or more comprehensive data which was not considered during the initial deployment. One interviewee noted \textit{``If you build a model on 10 year old data, then you’re not representing the current state of risks for certain disease. As a minimum, [recalibration] on new data would probably be more meaningful'' (P18).} In addition to reliability, continuous validation and improvement of other ethics principles may occur at run-time. System-level updates is necessary to address unethical issues. 

\begin{figure}[h!]
\begin{center}
\noindent
\fbox{%
\begin{minipage}{0.99\columnwidth}
\normalsize
    \textbf{Finding 12:} There is a strong desire for continuously monitoring and validating AI systems post deployment for responsible AI requirements but current MLOps practices provide limited guidance.
\end{minipage}}
\end{center} 
\end{figure}

\noindent \textbf{Traceability of artifacts.}
One approach often identified by the interviewees are related to traceability, provenance and reproducibility, which are useful to building trust in AI systems. It is necessary to track the use of an AI system and model provenance to improve explainability and accountability.
One interviewee mentioned \textit{``Things that I was on that had very – very strict rules about the provenance. So basically, every piece of code and every output had to go somewhere and have metadata tagged with it, so that if anyone wanted to audit what we did they could'' (P4).} It is well perceived that version control and immutable log are important for model provenance. One interviewee mentioned \textit{``When the system gets complex, you have to keep more evidence along the way. Version control, and the immutable log. You don’t want people to tamper this since after things went wrong'' (P2).} This improves both the trust and trustworthiness of AI systems. We found most of the interviewees used Git repository management software tools (e.g. GitHub or Bitbucket) for code version code. \textit{``Any software we are developing is in Bitbucket, internal configuration management system'' (P17).} However, an AI system usually involve co-evolution of data, model, code, and configurations. Thus, data/model/code/configuration co-versioning with model dependency specification is needed to ensure data provenance and traceability. If AI models are based on domain knowledge models, the underlying domain knowledge models need to be co-versioned with the AI models. There has been a lack of tools to use these traceability and provenance data to help with ethical AI concerns.



\begin{figure}[h!]
\begin{center}
\noindent
\fbox{%
\begin{minipage}{0.99\columnwidth}
\normalsize
    \textbf{Finding 13:} An AI system usually involve co-evolution of data, model, code, and configurations.  Data / model / code / configuration co-versioning with model dependency specification is needed to ensure data provenance and traceability.
\end{minipage}}
\end{center} 
\end{figure}


\begin{figure}
\centering
\includegraphics[width=0.5\columnwidth]{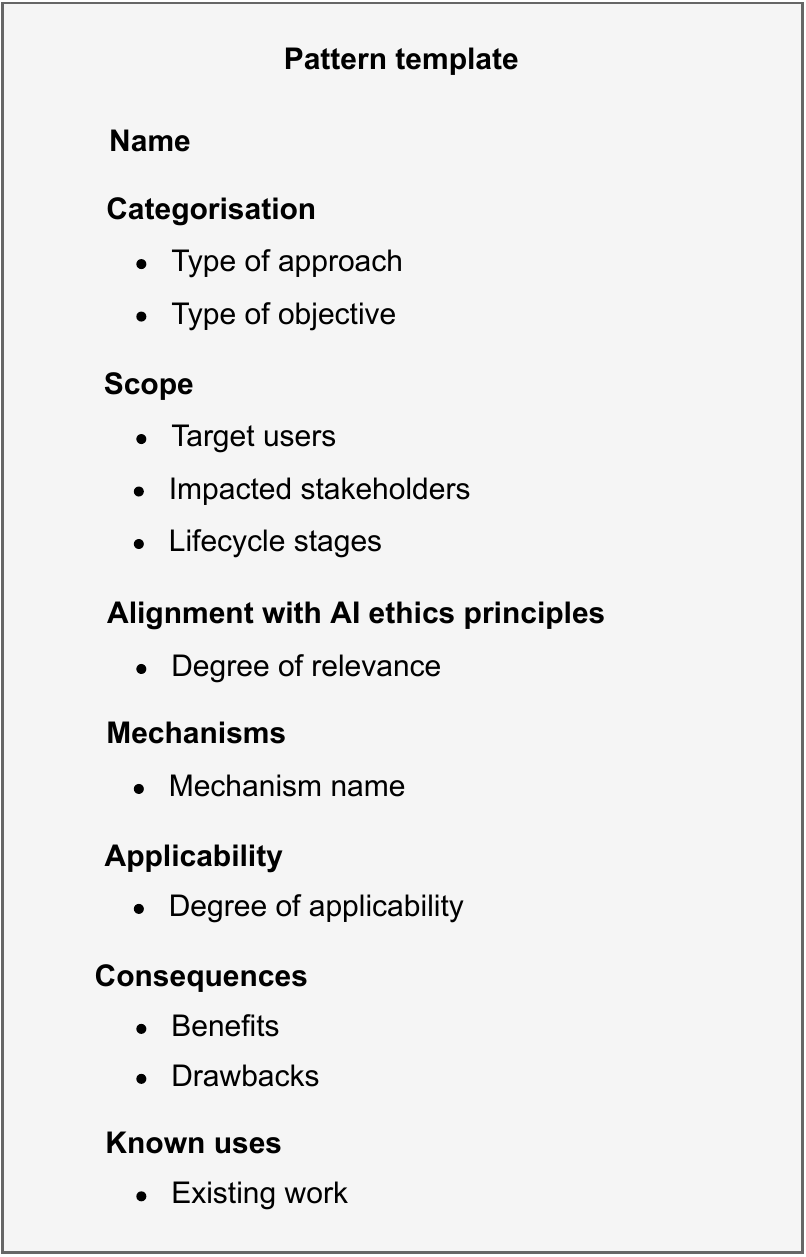}
\caption{Template of patterns for responsible AI.} \label{fig:template}
\vspace{-2ex}
\end{figure}

\begin{figure*}
\centering
\includegraphics[width=1.20\textwidth, angle=90]{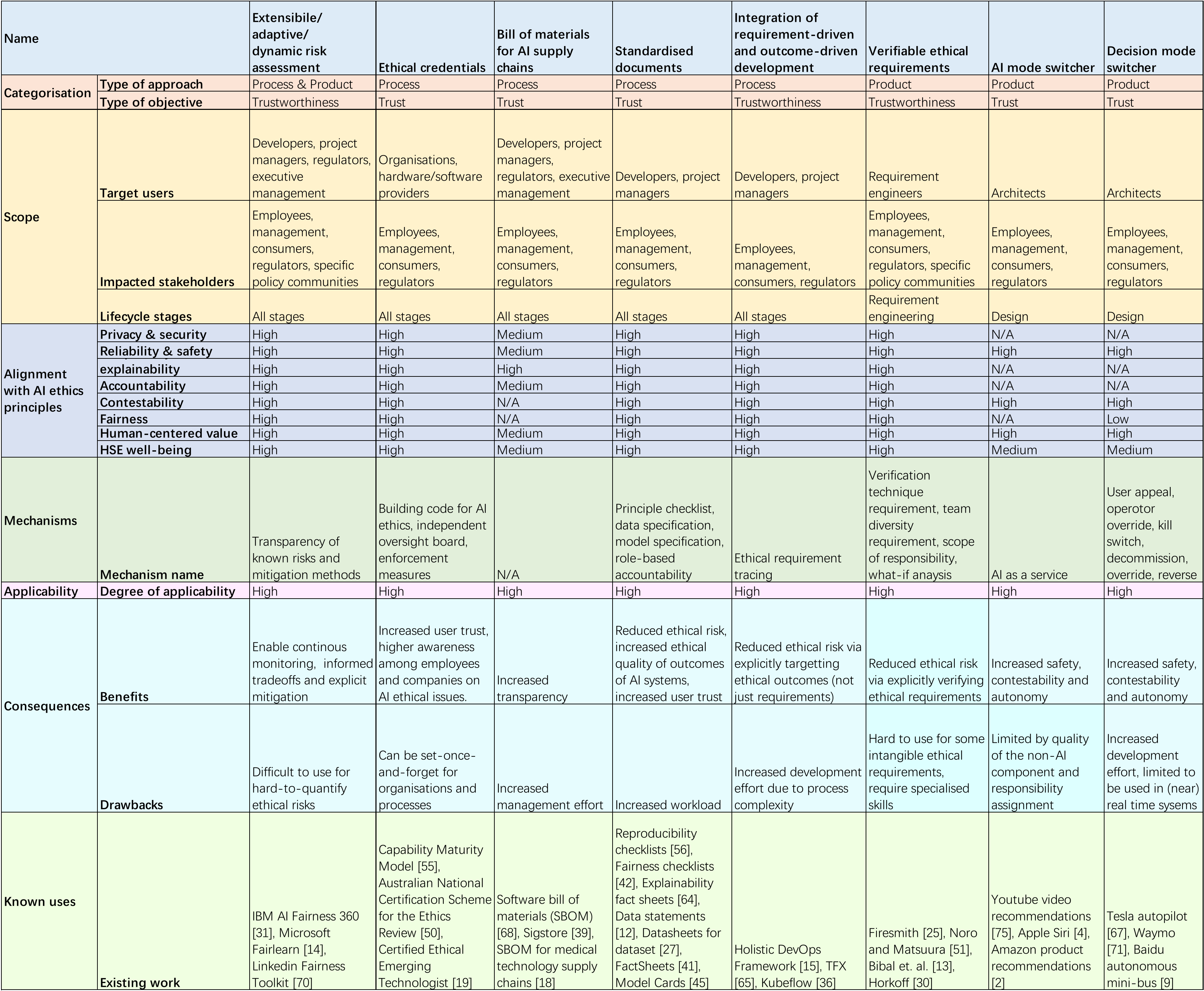}
\caption{Operationalised patterns for responsible AI --- part 1.} \label{fig:pattern1}
\end{figure*}

\begin{figure*}
\centering
\includegraphics[width=1.20\textwidth, angle=90]{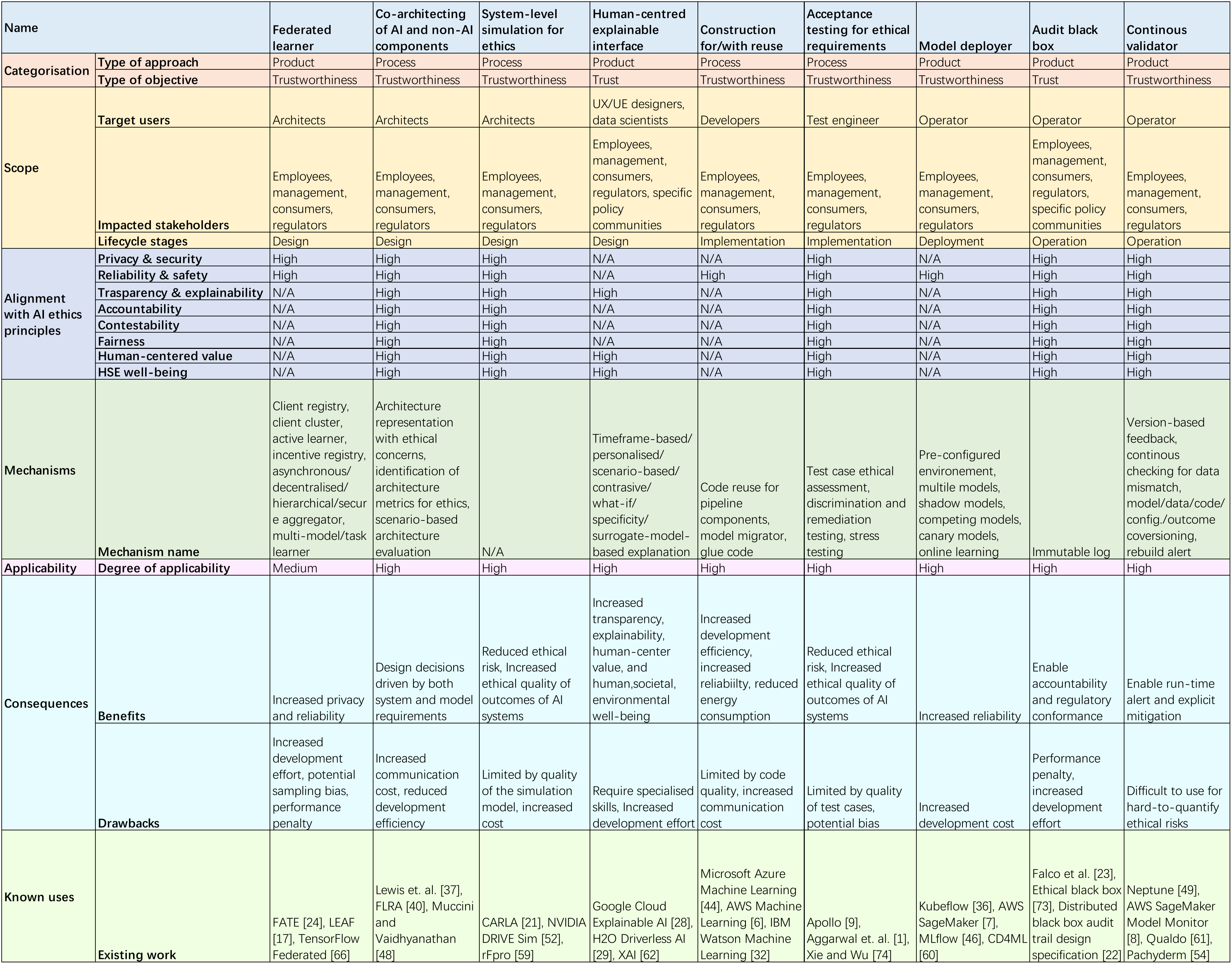}
\caption{Operationalised patterns for responsible AI --- part 2.} \label{fig:pattern2}
\end{figure*}

\section{Operationalised Patterns}
\label{mechanisms}
To operationalise responsible AI, as shown in Fig.~\ref{fig:template}, we define a pattern template which provides an integrated view of the the following aspects: categorisation, scope, alignment with AI ethics principles, mechanisms, applicability, consequences, and know uses. 
In Fig.~\ref{fig:pattern1}-\ref{fig:pattern2} we summarise a list of operationalised responsible AI assurance patterns
using the newly defined template based on the interview results, literature review, and existing software development practices. 

\section{Threats to Validity}
\label{validity}

\subsection{Internal Validity}
In our study, the interviewees were selected via ``call for participation'' emails
and recommendations within one organisation.
Although selection bias is always a concern when the interviewees are not randomly sampled,
the procedure partially alleviates the threat since the interviewers
have no contact with interviewees before the interviews.
Furthermore, given that our interviews include practitioners with various backgrounds, roles, and genders,
the threat has limited influence. 

We stopped our interviews when we achieved a saturation of findings after interviewing 21 persons.
To avoid the risk of missing information and interviewer subjectivity,
each interview included three interviewers with various research backgrounds.
The three interviewers worked together to ask questions and take notes during interviews.
This can aid in reducing the likelihood of subjective bias on whether the saturation of findings has been achieved,
as well as maximising the capture of as much relevant data as possible.

The operationalised patterns we recommended may not cover all the existing solutions for some of the development stages and AI ethics principles, e.g. testing, technologies for reliability, as they have been well studied in significant literature. The emphasise of our work is mainly on the stages and ethics principles that are still under explored and hard to be operationalised, e.g. requirement engineering, architecture design, and DevOps.

\subsection{External Validity}
This study is conducted within one organisation, which may introduce a threat to external validity.
While we recognise that having more organisations would be desirable,
we believe our study is generalisable to most AI system development teams.
All the interviewees are from a national science agency with teams working on multiple areas
serving various customers, and having various products/projects and cultures.
We acknowledge that the opinions provided by our interviewees may not be representative of the whole community.
To reduce this threat, we ensured that our interviewees hold various roles and have various levels of expertise.
We believe that their opinions and comments uncovered various insights into the challenges developers are facing
in dealing with responsible AI issues during development. 

\section{Related Work}
\label{related_work}
The challenge of responsible AI has gathered considerable attention~\cite{zhang21}.
Nearly 100 high-level principles and guidelines for AI ethics have been issued by governments, organisations, and companies~\cite{jobin2019global}.
A degree of consensus around high-level principles has been achieved~\cite{fjeld2020principled}.
Certain AI ethics principles, such as privacy \& security, reliability \& safety, and fairness,
can be considered as software quality attributes.
Security, reliability and safety are well-studied in the dependability research community~\cite{dependability}
and can be specified as non-functional requirements to be considered in the development.
There are reusable design methods (e.g., patterns) that could be applied to address these principles~\cite{bass2003software}.
Although privacy is not a standard software quality attribute~\cite{iso2011iec25010},
it has been increasingly taken into consideration as an important requirement of a software system
in the design to conform with regulation, e.g., General Data Protection Regulation (GDPR)~\cite{gdpr}.
Patterns have been summarised to address privacy concerns and realise privacy-by-design~\cite{privacy_patterns}. 
Fairness is a quality attribute that AI developers should consider throughout the AI system lifecycle.
Many methods and tools have been introduced into AI pipeline to achieve fairness more at model-level
rather than system-level \cite{Mehrabi21fairness}, such as IBM's AI Fairness 360~\cite{IBMAIFairness360},
Microsoft Fairlearn~\cite{Fairlearn}, and Linkedin Fairness Toolkit~\cite{Linkedin_Fairness_Toolkit}. 

Human, societal and environmental wellbeing, as well as human centered values can be treated as functional requirements.
However, there is lack of work on the operationalisation of these two principles.
For example, for the human-centered values principle,
which values are considered and how can these values be designed for,
implemented and tracked in an AI system.
Risk mitigation mechanisms and the existing approaches on operationalising human value in software
can be applied to achieve these two principles in AI systems \cite{Whittle18, Whittle19}.
Transparency \& explainability, contestability, and accountability
can be viewed as meta-level governance-related meta-level functional requirements.
New design and process patterns are needed to fulfil these principles, particularly from a governance perspective.

Overall, AI ethics principles need to be operationalised in the form of concrete practices
that are usable by AI developers when developing AI systems.
Although OECD provides a tool framework for trustworthy AI~\cite{oecd},
the framework largely contains categorised but disjointed software tools and guidelines,
lacking process-related linkages and the trust side in addition to trustworthiness.
Thus, an operationalised guidance for developers is required throughout the entire lifecycle of AI systems.   

\section{Conclusion}
\label{conclusion}
AI ethics principles are typically high-level and do not provide concrete guidance to developers on how to develop AI systems responsibly. In this study, we first perform an empirical study to understand the practitioners' perceptions on AI ethics principles and their implementation. We then suggest a list of patterns to provide a concrete, operationalised guidance that are usable by AI developers to develop responsible AI systems.

\input{main.bbl}


\end{document}